\documentclass[letterpaper, 10 pt, conference]{ieeeconf}  

\IEEEoverridecommandlockouts                              

\usepackage{graphicx}
\usepackage{float}  
\usepackage{color}
\usepackage{amsmath}
\usepackage{booktabs}
\usepackage[colorlinks=true, linkcolor=blue]{hyperref}
\usepackage{cleveref}
\usepackage{multirow}
\usepackage{comment}
\usepackage{amssymb}
\usepackage{pifont}

\title{\LARGE \bf
CoT-VLM4Tar: Chain-of-Thought Guided Vision-Language Models for Traffic Anomaly Resolution
}

\author{Tianchi Ren$^{*, 1}$, Haibo Hu$^{*, 2}$, Jiacheng Zuo$^{3}$, Xinhong Chen$^{4}$, Jianping Wang$^{5}$,\\ Chun Jason Xue$^{6}$, Jen-Ming Wu$^{7}$, Nan Guan$^{\dag, 8}$
\thanks{*Equal contribution \qquad \dag corresponding author}
\thanks{$^{1}$City University of Hong Kong Material Science Institute. 
        {\tt\small rentc2003@stu.xjtu.edu.cn}}%
\thanks{$^{2, 4, 5, 7}$Department of Computer Science, City University of Hong Kong.
        {\tt\small haibohu2-c@my.cityu.edu.hk,}
        {\tt\small \{xinhong.chen, jianwang, nanguan\}@cityu.edu.hk}}%
\thanks{$^{3}$Department of Computer Science, Soochow University.
        {\tt\small 20235227060@stu.suda.edu.cn}}%
\thanks{$^{3}$Department of Computer Science, Mohamed bin Zayed University of Artificial Intelligence.
        {\tt\small jason.xue@mbzuai.ac.ae}}%
\thanks{$^{3}$Honhai Research Institute.
        {\tt\small jen-ming.wu@foxconn.com}}%
}

\begin{document}

\maketitle
\thispagestyle{empty}
\pagestyle{empty}

\begin{abstract}
With the acceleration of urbanization, modern urban traffic systems are becoming increasingly complex, leading to frequent traffic anomalies. These anomalies encompass not only common traffic jams but also more challenging issues such as phantom traffic jams, intersection deadlocks, and accident liability analysis, which severely impact traffic flow, vehicular safety, and overall transportation efficiency. Currently, existing solutions primarily rely on manual intervention by traffic police or artificial intelligence-based detection systems. However, these methods often suffer from response delays and inconsistent management due to inadequate resources, while AI detection systems, despite enhancing efficiency to some extent, still struggle to handle complex traffic anomalies in a real-time and precise manner. To address these issues, we propose CoT-VLM4Tar:  (Chain of Thought Visual-Language Model for Traffic Anomaly Resolution), this innovative approach introduces a new chain-of-thought to guide the VLM in analyzing, reasoning, and generating solutions for traffic anomalies with greater reasonable and effective solution, and to evaluate the performance and effectiveness of our method, we developed a closed-loop testing framework based on the CARLA simulator. Furthermore, to ensure seamless integration of the solutions generated by the VLM with the CARLA simulator, we implement an itegration module that converts these solutions into executable commands. Our results demonstrate the effectiveness of VLM in the resolution of real-time traffic anomalies, providing a proof-of-concept for its integration into autonomous traffic management systems. 
\end{abstract}

\section{INTRODUCTION}
Traffic anomalies, such as ghost jams~\cite{ghost} and traffic accidents~\cite{accident}, have become a significant challenge in modern urban transportation systems~\cite{c1}. These irregularities not only hinder traffic flow but also contribute to increased congestion, travel time, and environmental pollution~\cite{c2,c3}. The growing complexity of traffic management in cities with high vehicle density and limited resources exacerbates these issues~\cite{c5}. For example, traffic congestion in Chittagong's port and industrial areas causes daily economic losses of \$2.01 million, high stress levels, and operational inefficiencies~\cite{mjl}.

Currently, traffic anomalies are primarily addressed by the intervention of traffic police at the scene ~\cite{tp1,tp2}or through AI-based detection systems~\cite{c6} that trigger human intervention at control centers. These solutions typically involve manual assessment, coordination with various stakeholders, and the deployment of personnel to manage the situation~\cite{tp1,tp2}. Though this approach is effective in general, it is highly time- and labor-consuming since traffic anomalies can occur anytime and anywhere, often worsening before law enforcement arrives, leading to greater delays~\cite{tc4a}. Additionally, relying on human operators requires continuous recruitment and training, which is costly and impractical given limited resources.

Addressing these limitations requires reducing the time and labor demands of current solutions. Human intervention is not always available, and even when present, officers may struggle to quickly assess complex, rapidly changing traffic scenarios, such as multi-vehicle accidents or sudden blockages. These delays worsen congestion and increase the risk of further incidents. 
Moreover, continuous recruitment and training of skilled personnel are costly and unsustainable, especially in urban areas with frequent anomalies. An automated, real-time response system is essential to alleviate the burden on human operators and ensure faster, more efficient traffic anomaly management.

The recent rise of VLMs has inspired new perspectives. VLMs integrate visual and textual data, enabling contextual understanding and reasoning over complex traffic scenes~\cite{c7,c8,vlm4cap}. By processing multimodal inputs from traffic cameras, sensors, and drones, VLMs can interpret real-time anomalies and generate actionable solutions~\cite{vlm3,vlm1,c10}. Their ability to reason over dynamic environments makes them well-suited for autonomous traffic management, improving response efficiency in scenarios like congestion, intersection deadlocks, and accident analysis~\cite{vlm5,c11,c12}. Therefore, utilizing VLMs to address traffic anomaly resolution time, enable rapid response, and replace manual intervention presents an effective solution.

Leveraging the powerful reasoning and analytical capabilities of VLMs to rapidly respond to and directly engage in traffic anomaly scenarios presents an effective solution to the current handling of traffic anomalies. Nevertheless, this approach faces several challenges: (1) the replication of diverse traffic anomaly scenarios in a controlled environment; (2) while VLMs possess significant capabilities, they cannot be directly applied to traffic anomaly resolution without adaptation to the specific nuances of traffic management; and (3) validating the effectiveness of VLM-generated solutions remains a key challenge, requiring robust methods for assessing the model's real-world impact and ensuring the reliability of its decisions. Addressing these challenges is critical for successfully integrating VLMs into autonomous traffic anomaly management systems.

In this paper, we propose CoT-VLM4Tar (VLM guided by Chain-of-Thought for traffic anomaly resolution), A novel traffic anomaly chain-of-thought is introduced to guide the VLM in analyzing, reasoning, and ultimately generating solutions for these anomalies. We also observe that there has been little effective validation of the VLM's ability to analyze and generate solutions for real-world traffic scenarios. To address this, we employ the CARLA simulator to recreate traffic scenarios and establish communication between CARLA and the VLM. To ensure seamless integration, we standardized the output format of the VLM, allowing us to develop an integrated module that converts the proposed solutions into executable commands within CARLA. This creates a dynamic processing loop between CARLA scenarios and the VLM, enabling closed-loop testing.

Our specific contributions are as follows:
\begin{itemize}
\item Closed-Loop Testing Framework: We propose a closed-loop framework using VLMs in CARLA to simulate and address traffic anomalies like ghost traffic jams, intersection deadlocks, and accidents, showcasing their potential in real-time traffic management.

\item Traffic Anomaly Chain-of-Thought: We introduce a novel Chain-of-Thought approach to guide VLMs in analyzing and resolving traffic anomalies with improved accuracy and efficiency.

\item VLM Effectiveness in Traffic Anomalies: Our results demonstrate VLMs’ capability in real-time traffic anomaly resolution, highlighting their potential for autonomous traffic management and flow optimization.
\end{itemize}

\section{Related Work}
\begin{figure*}[!t]
    \centering
    \includegraphics[width=1\textwidth]{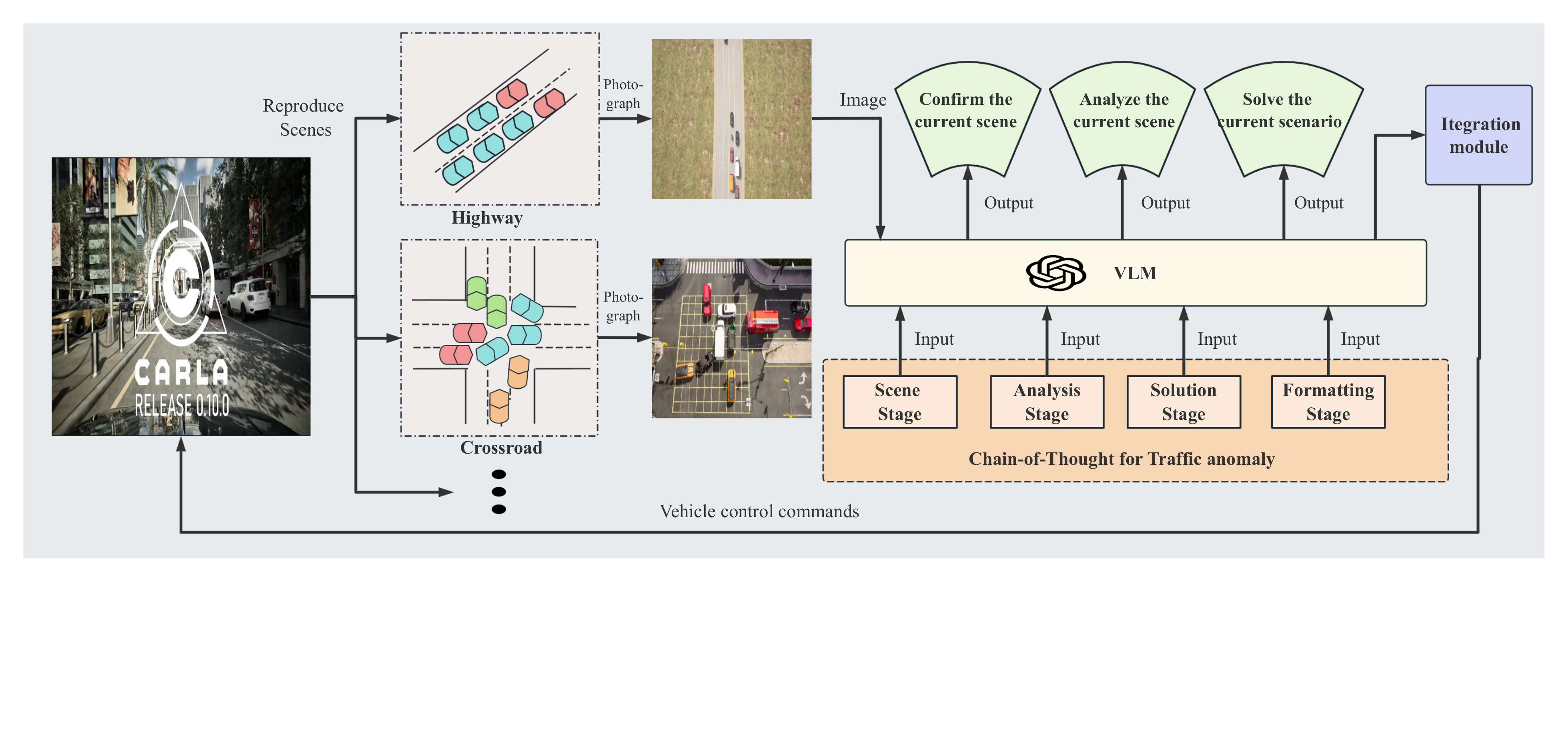}
    \vspace{-80pt}
    \caption{Overview of the Closed-Loop Testing Framework for Traffic Anomaly Resolution Using VLMs.}
    \vspace{-10pt}
    \label{fig:overview}
    \end{figure*}
\subsection{Driving Simulator in Traffic}
Driving simulators based on 3D modeling have been widely applied in the fields of traffic and autonomous driving, such as Airsim~\cite{airsim}, LGSVL Simulator~\cite{lgsimulator}, and CARLA~\cite{carla}. Among these, CARLA stands out due to its powerful capabilities and open-source nature. It enables the creation of diverse traffic scenarios, including city layouts, numerous vehicle models, buildings, pedestrians, traffic signs, and more. CARLA also features a robust perception system, providing detailed data on GPS coordinates, speed, acceleration, as well as information on collisions and other violations. Various environmental conditions, such as weather and time of day, can be specified. As a result, many researchers choose CARLA for traffic-related experiments. For example, Hartwich et al. used a driving simulator to examine the impact of autonomous driving and driving style familiarity on driving comfort, enjoyment, and system acceptance~\cite{simulator4comfortable}. X. Liang simulated traffic signal systems in a driving simulator to experiment with the use of AI for improving traffic efficiency~\cite{si4light}. A. Olia utilized simulators to analyze the effects of mixed autonomous and human-driven vehicles on traffic flow~\cite{si4mix}. Therefore, employing a driving simulator to recreate traffic anomaly scenarios is an effective approach.
\subsection{VLM in Traffic}
The application of Vision-Language Models (VLMs) in traffic analysis is gaining traction. In~\cite{vlm1}, models like VideoLLaMA-2 and GPT-4o were evaluated on real-world and synthetic traffic videos to answer complex queries on traffic conditions and events. Shoman et al.~\cite{vlm2} integrated object detection, tracking, and language generation to enhance traffic event analysis. Jain et al.~\cite{vlm3} fused VLMs with multi-sensor data to improve understanding of traffic dynamics. TrafficVLM~\cite{vlm4} was introduced for dense video captioning, generating detailed descriptions of traffic scenarios. Additionally, iLLM traffic signal control (TSC)~\cite{vlm5} employed a two-step decision process combining reinforcement learning and contextual reasoning. However, no existing work has applied VLMs to traffic anomaly detection.
\subsection{Chain of Thought}
LLMs can perform complex reasoning tasks by generating intermediate reasoning steps through a process known as thought chaining (LC)~\cite{thougt1, llm1}. This concept has inspired researchers to utilize self-generation principles for demonstrations. Specifically, Zelikman et al.~\cite{thougt2} demonstrated the practicality of using LLM-based generation principles. They prompted GPT-J~\cite{thougt3} to generate principles and then selected those that led to the correct answers. Zero-shot LC prompting employs simple instructions to guide step-by-step reasoning before arriving at a solution. LLMs exhibit reasonable zero-shot reasoning capabilities, with the generated output essentially reflecting LC-based reasoning~\cite{thougt4}. In our approach, we adopt this method with VLMs, starting with simple scene classification, followed by targeted analysis based on the classification, and ultimately providing a solution derived from the analysis.

\section{METHODOLOGY}
This section provides a detailed explanation of the CoT-VLM4Tar, as shown in Figure~\ref{fig:overview}. we developed the framework to utilize VLMs for traffic anomaly resolution in a CARLA simulator environment. The process begins with the reproduction of common traffic anomaly scenarios using CARLA. These scenarios are then sampled for analysis by the VLM, which, guided by a traffic anomaly chain-of-thought, to evaluate the situation and provide solutions, such as adjusting the speed of a vehicle. The VLM outputs a standard solution format, which is then converted into executable CARLA commands via the API command conversion module, completing the loop. This system allows for the real-time evaluation of VLM-driven traffic anomaly resolution and demonstrates the model's effectiveness in handling complex traffic situations.

\subsection{Traffic Anomaly Replay in CARLA}
\begin{figure}[thpb]
      \centering
      \vspace{-10pt}
      \includegraphics[width=0.48\textwidth]{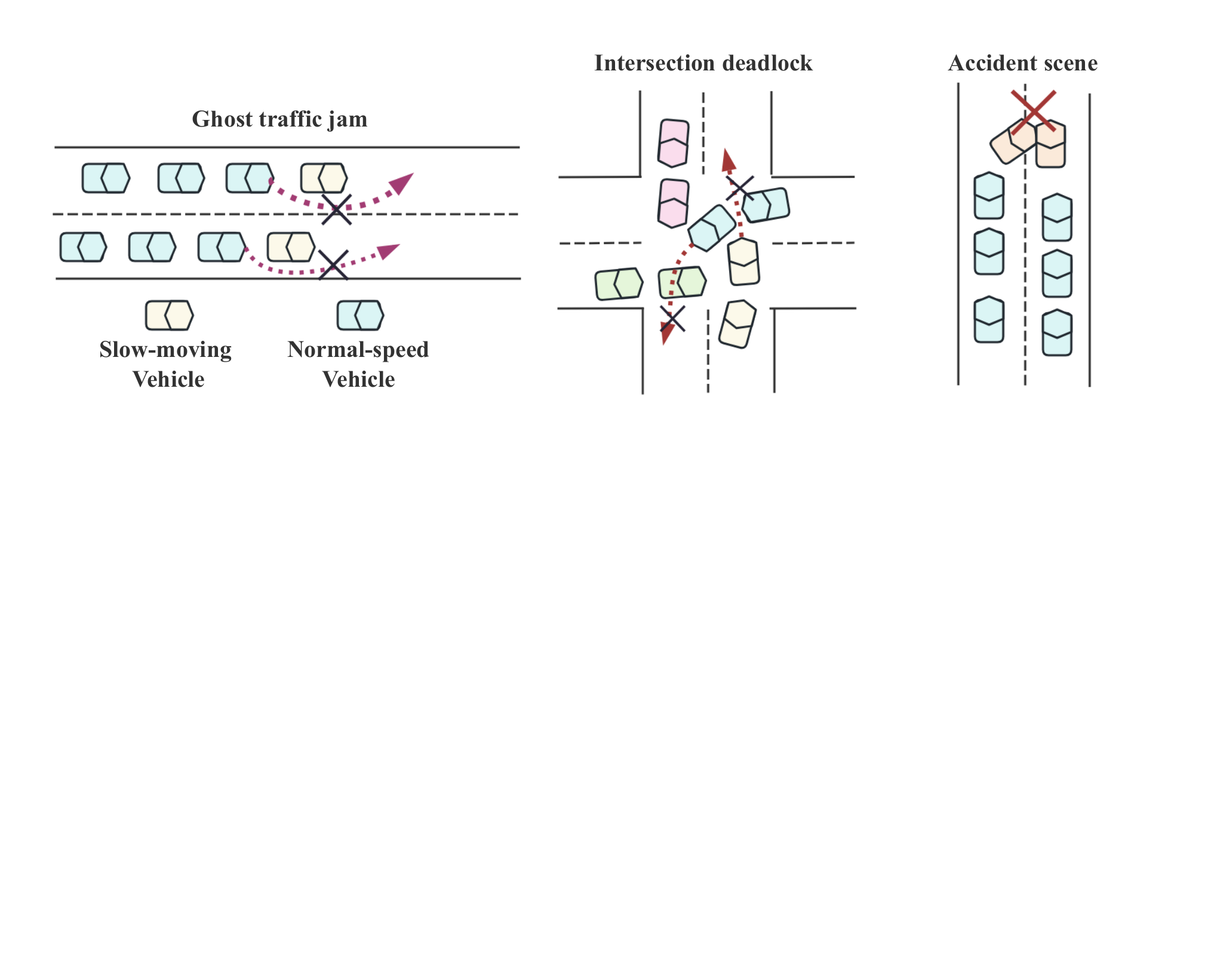}
      \vspace{-130pt}
      \caption{Visualization of three scenarios}
      \vspace{-10pt} 
      \label{fig2}
    \end{figure}

In this study, we employ the CARLA simulator to recreate various traffic anomaly scenarios, providing a controlled environment for testing the effectiveness of VLMs in real-time traffic management. We selected three specific scenarios—ghost traffic jams, intersection deadlocks, and collision events—for detailed analysis and testing. These scenarios were chosen because they represent common traffic anomalies that significantly impact traffic flow and pose challenges in autonomous traffic management, as shown in Figure~\ref{fig2}.

\textbf{Ghost Traffic Jams:} Often caused by inefficient vehicle behavior, such as slow-moving vehicles blocking overtaking, leading to bottlenecks.

\textbf{Intersection Deadlocks:} A typical issue where conflicting vehicle movements prevent vehicles from clearing an intersection, resulting in gridlock.

\textbf{Collision Events:} Common in traffic systems, accidents disrupt flow and require quick resolution to mitigate congestion.

By recreating these scenarios, we can assess how VLMs handle common, real-world traffic anomalies, providing a robust test for their effectiveness.

For the vehicle generation process, we rely on CARLA’s native API to instantiate a range of vehicle types within the simulator. Vehicles are placed strategically in the environment to replicate particular traffic anomalies. For instance, in the case of ghost traffic jams, we simulate two slow-moving parallel vehicles that impede faster vehicles from overtaking, leading to a traffic bottleneck. Similarly, we recreate intersection deadlocks by positioning vehicles in such a way that multiple cars are unable to navigate through the intersection due to conflicting movement patterns. Other scenarios, such as collision events, are also simulated by adjusting vehicle speeds and positions to emulate real-world accident conditions. 
This method allows us to generate diverse and realistic traffic anomaly scenarios that serve as test cases for evaluating the performance of the VLMs. 

\subsection{Chain-Of-Thought for Traffic Anomaly}
\begin{figure}[thpb]
      \centering
    \includegraphics[width=0.5\textwidth]{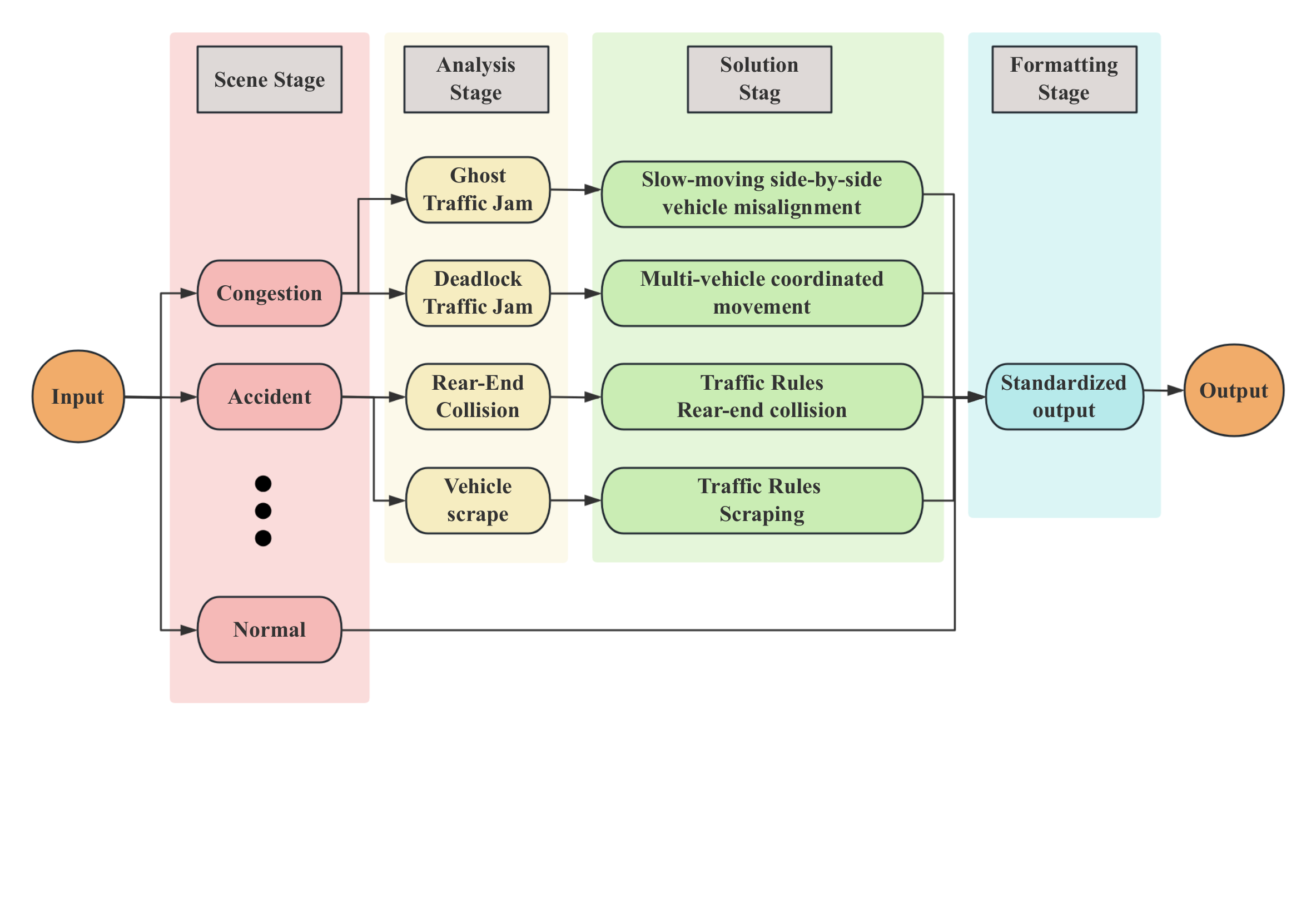}
    \vspace{-55pt} 
      \caption{Chain-Of-Thought for Traffic Anomaly}
      \label{cot}
    \end{figure}

The effectiveness of Vision-Language Models (VLMs) in resolving traffic anomalies is greatly enhanced by guiding the model through a structured multi-step reasoning process. While VLMs are capable of generalizing across different traffic scenarios, they tend to offer broad interpretations that might not directly generate the specific actions needed to resolve traffic issues. This is similar to findings in previous models like LLaVA-CoT~\cite{llm-cot}, where structured guidance was required to enable the model to incrementally achieve the desired outcomes. To effectively handle traffic anomalies, we propose a four-stage CoT that helps guide the VLM through a precise reasoning process, as show in Fig~\ref{cot}.

\textbf{Scene Stage (Classification):} The first step in the Cot is to classify the traffic scene based on the input from CARLA. The VLM assesses the scenario to determine if it depicts normal traffic flow, congestion, a ghost traffic jam, an intersection deadlock, or an accident site. This classification serves as the foundation for the subsequent reasoning process. By identifying the type of traffic anomaly, the model sets the context for its analysis and narrows the focus to relevant factors, ensuring more accurate interpretations.

In the CoT framework, this stage can be viewed as the model asking itself: “What type of anomaly is occurring in this scene?” The answer to this question then guides the analysis.

\textbf{Analysis Stage (Reasoning):} Once the scene is classified, the VLM enters the analysis phase, where it reasons through the dynamics of the traffic situation. Here, the model breaks down the scene to examine key factors such as vehicle positioning, speed differences, and interaction patterns. It aims to understand the underlying causes of the anomaly. For example, in the case of a ghost traffic jam, the model would analyze how slow-moving vehicles side-by-side contribute to the congestion.

In this stage, the CoT approach is central: the model employs a step-by-step reasoning process, such as “Given the classification, what dynamics are causing this specific type of anomaly?” The analysis allows the model to pinpoint the contributing factors, which sets the stage for devising an appropriate solution.

\textbf{Solution Stage (Actionable Interventions):} After analyzing the scene, the VLM proposes a solution tailored to the identified anomaly. This solution is derived from the previous reasoning process, ensuring that the interventions are contextually relevant. For example, for a ghost traffic jam, the model may suggest that vehicles adjust their positioning to alleviate the bottleneck. In the case of an intersection deadlock, the model might recommend changing traffic signal timings or rerouting vehicles.

In the CoT framework, this stage is driven by the model asking itself: “What is the best course of action to address the identified issue?” By following this chain of reasoning, the model ensures that the proposed solution aligns with the context and complexity of the traffic anomaly.

\textbf{Formatting Stage (Execution):} Finally, the solution must be formatted in a standardized way that aligns with CARLA's command execution framework. This ensures that the model's proposed actions are translated into executable instructions. The formatting stage transforms the VLM’s reasoning into a unified format, which can then be processed by the integration module in a rule-based manner.

\textbf{Closing the Loop with CoT:}
Through this four-stage process—Scene, Analysis, Solution, and Formatting—we guide VLMs to resolve traffic anomalies by structuring their reasoning and actions. The Chain-of-Thought method ensures that VLMs approach the task step-by-step, refining their understanding at each stage, ultimately leading to more accurate, practical, and executable solutions. This structured CoT not only enhances the model's ability to handle complex traffic scenarios but also ensures that its outputs can be directly tested and implemented in real-time simulations.

\subsection{CARLA and VLM Integration Module}
\begin{figure}[thpb]
        \vspace{-10pt}
      \centering
    \includegraphics[width=0.5\textwidth]{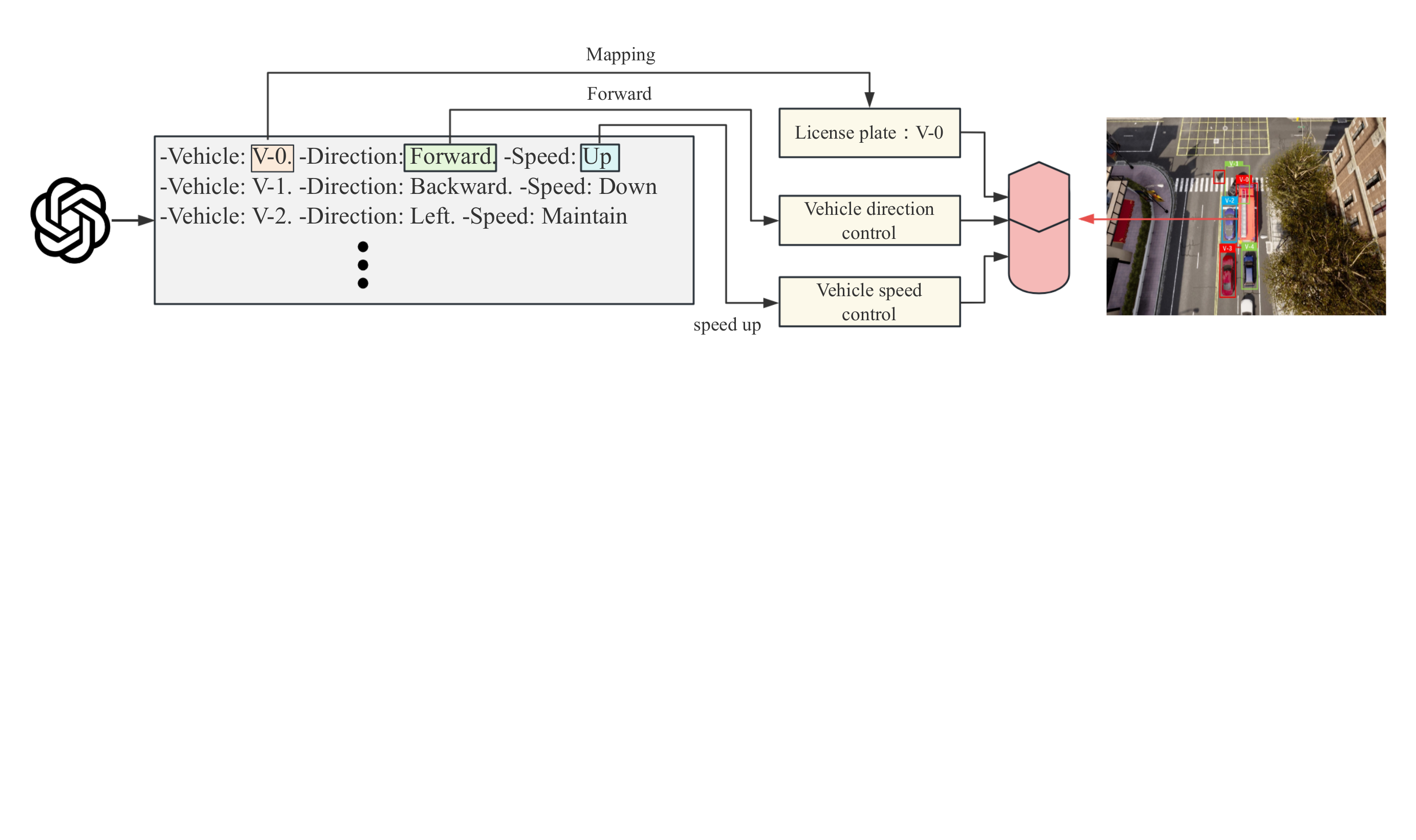}
    \vspace{-100pt} 
      \caption{The process of converting VLM output results into executable commands for Carla}
      \vspace{-5pt} 
      \label{fig4}
    \end{figure}

The final stage of our methodology involves translating the structured solutions generated by the VLM into executable actions within the CARLA simulator. After the VLM produces a solution in the standardized output format, we extract specific details such as the vehicle IDs and the corresponding actions to be performed. This step is crucial for ensuring that the proposed solutions align with the real-time control capabilities of the CARLA simulator. First, the output from the VLM is parsed to identify key parameters, including the target vehicle(s), the desired behaviors (e.g., speed adjustments, lane changes), and any additional instructions related to the environment, such as traffic signal modifications or obstacle avoidance. For example, if the VLM suggests accelerating a specific vehicle to resolve a ghost traffic jam, the vehicle ID and the required speed increase are extracted, as show in Fig~\ref{fig4}.
    
Next, the extracted parameters are mapped to the corresponding data structures and commands within the CARLA API. CARLA’s API provides a comprehensive set of functions to interact with the simulation, including vehicle control (e.g., throttle, brake, steering), traffic management, and environmental modifications. By leveraging these capabilities, we translate the VLM’s solution into a set of actionable commands that are compatible with CARLA’s real-time simulation environment.

Finally, these commands are sent to the CARLA simulator, where they are executed to modify the state of the traffic scenario. The simulation environment is then updated based on the model’s actions, allowing us to observe the impact of the proposed solutions on the traffic flow. This closed-loop system enables real-time validation of the VLM's effectiveness in handling complex traffic anomalies, offering insights into the practical application of AI-driven traffic management.

\section{EXPERIMENT}

To further validate the effectiveness of TRA-VLM, we conducted experiments on three traffic anomaly scenarios: ghost traffic jam caused by slow-moving side-by-side vehicles, intersection deadlock, and accident responsibility allocation. The experiments were designed to evaluate how well TRA-VLM can analyze and generate solutions for these complex traffic issues, with the goal of improving traffic flow and decision-making in autonomous systems.
\subsection{Tools and Setup}
    \begin{table}[t]
    \vspace{10pt}
    \centering
    \caption{The validity test results of the VLMs are for the four-stage Chain-of-Thought}
    \label{table:vlms}
    \setlength{\tabcolsep}{0.2pt} 
    \begin{tabular}{|c|c|c|c|c|}
        \hline
        VLM Name &  Scene  Stage & 
                    Analysis Stage& 
                   Solution Stage  & 
                   Formatting Stage \\
        \hline
        MiniCPM14b &\checkmark &\ding{55} &\ding{55} &\ding{55}\\
        \hline
        VILA40b &\checkmark & \ding{55} & \ding{55} &\checkmark\\
        \hline
        ChatGPT-4o &\checkmark &\checkmark &\checkmark &\checkmark\\
        \hline
    \end{tabular}
    \end{table}  
For our simulations, we utilized CARLA version 0.9.15, which is widely recognized for its robustness and flexibility in autonomous driving research. This version of CARLA allows for customization of lanes and vehicles, and offers a comprehensive API that enables the effective creation of autonomous driving vehicles, facilitating the implementation of traffic anomaly scenarios. Additionally, the version provides perception data from V2X or drone perspectives, which is beneficial for replicating real-world scenarios and enabling rapid emergency responses.

We tested multiple VLMs, including GPT-4o, MiniCPM14b and VILA40b, for their ability to process and generate solutions based on the traffic data provided by CARLA, as shown in Table~\ref{table:vlms}. MiniCPM14b and VILA40b are unable to complete our tasks, particularly in the Analysis Stage and the Solution Stage, where they sometimes provide incorrect or irrelevant responses. In contrast, the ChatGPT-4o is set as our final model due to its superior performance in handling multimodal input and generating coherent, contextually relevant solutions. ChatGPT-4o's ability to integrate visual and language information efficiently makes it particularly well suited for analyzing traffic anomalies in real-time.

\subsection{Result}
\begin{figure}[h]
      \centering
    \includegraphics[width=0.5\textwidth,height=0.55\textwidth, keepaspectratio]{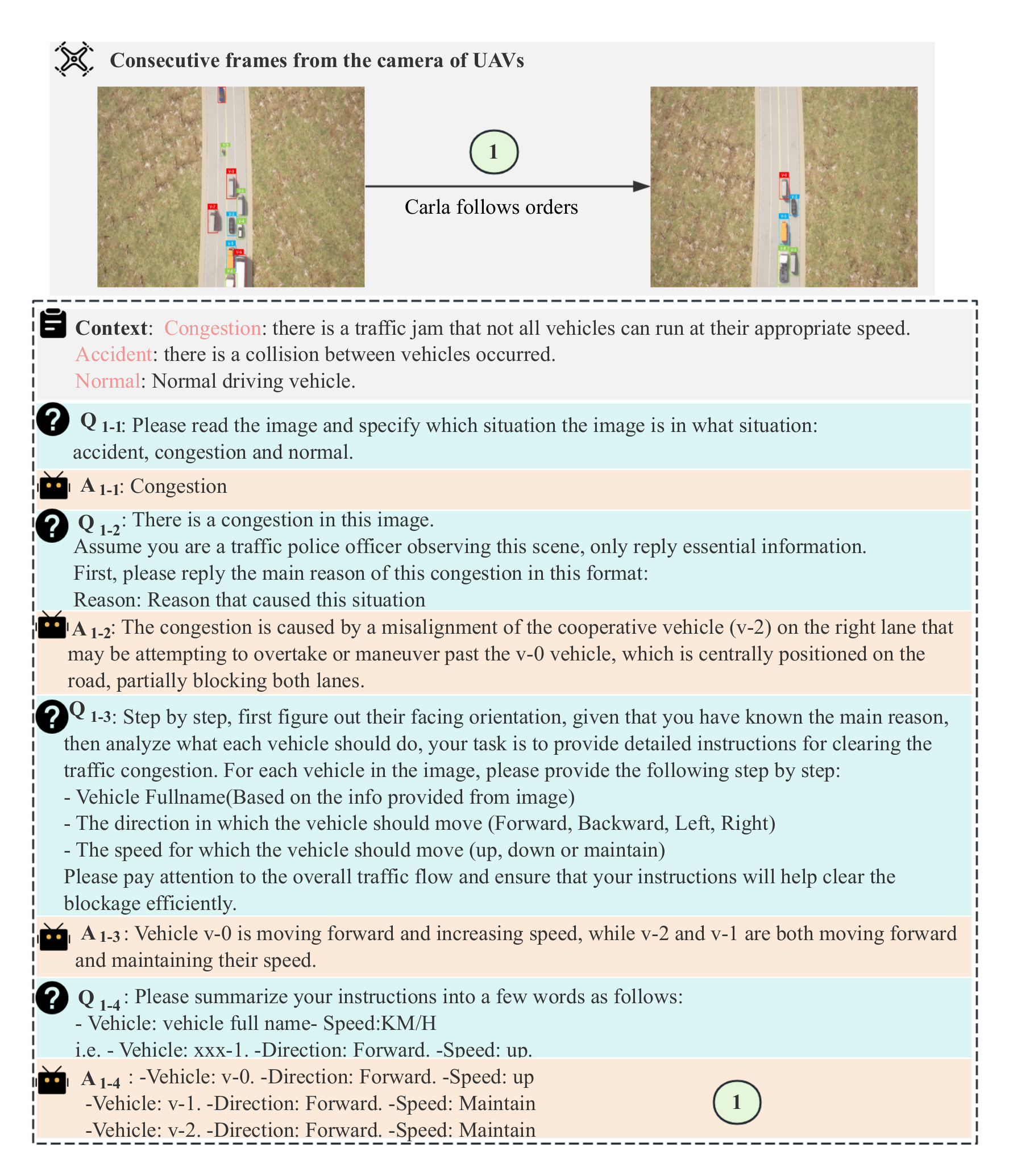}
    \vspace{-10pt} 
      \caption{the process of VLM handling phantom traffic jams}
    \vspace{-10pt} 
      \label{fig5}
    \end{figure}

\textbf{Ghost Traffic Jam from Slow Side-by-Side Vehicles:} This scenario simulates a ghost traffic jam caused by two slow-moving vehicles blocking adjacent lanes, leading to congestion. In the CARLA simulator, vehicles v-2 and v-0 travel slowly, obstructing the leftmost lanes and causing a traffic buildup. The VLM analyzes visual input, including sensor data, and identifies the congestion as a misalignment issue where v-2 attempts to overtake v-0, which is partially blocking both lanes, as shown in Fig~\ref{fig5}.
The VLM generates a solution by providing detailed instructions for each vehicle to clear the traffic bottleneck. For example, it instructs v-0 to move forward and increase speed, while v-2 should continue moving forward and maintain speed. Additionally, v-1 is instructed to move forward and increase speed to help optimize the overall traffic flow.  The effectiveness of the proposed solution is evaluated based on the reduction in travel time and the overall improvement in traffic speed.

\begin{figure}[h]
      \centering
    \includegraphics[width=0.5\textwidth, height=0.55\textwidth, keepaspectratio]{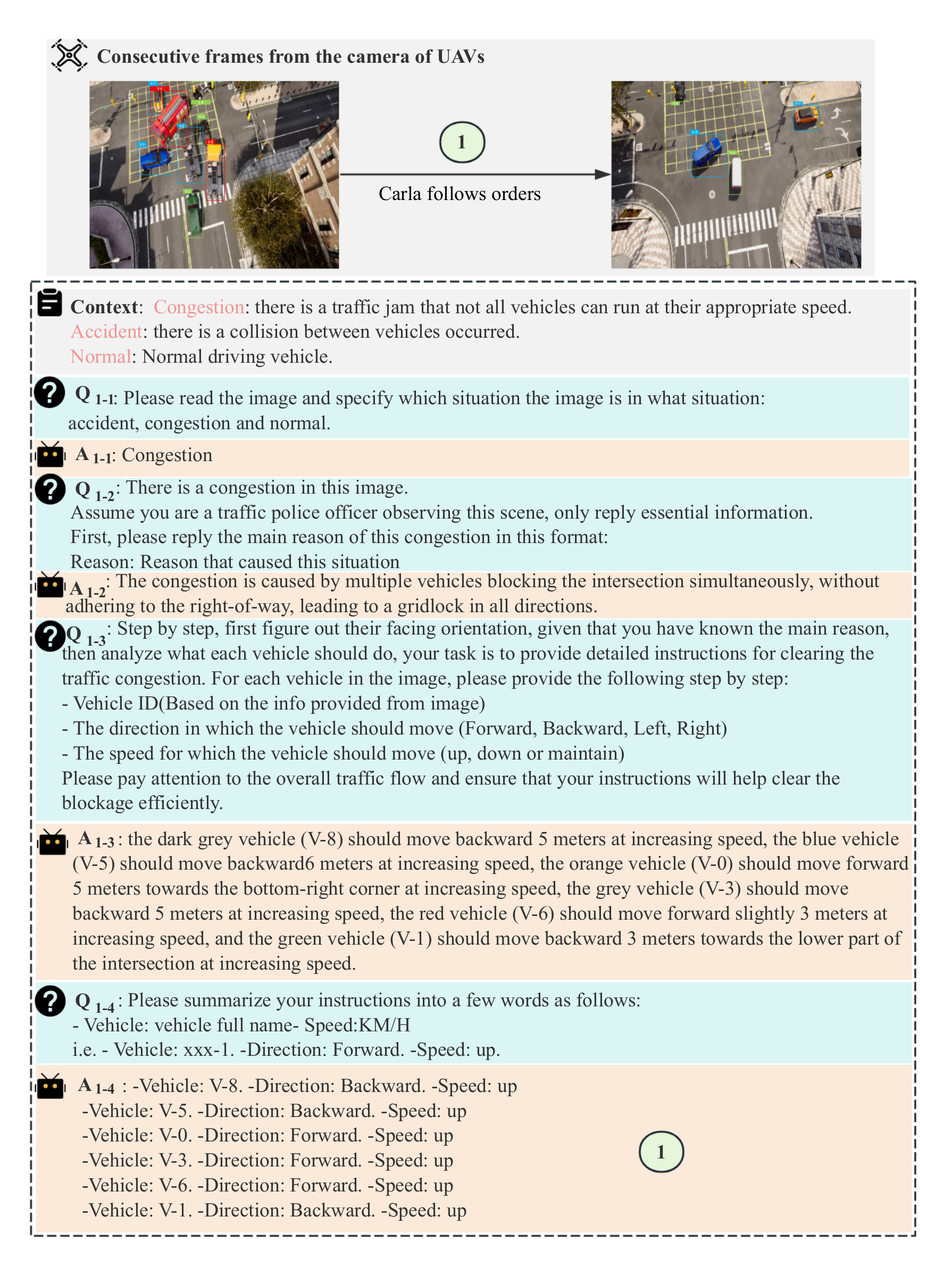}
    \vspace{-10pt} 
      \caption{the process of VLM handling Intersection Deadlock}
    \vspace{-10pt} 
      \label{fig6}
    \end{figure}

\textbf{Intersection Deadlock:} In the second experiment, we simulate an intersection deadlock scenario where multiple vehicles from different directions arrive at an intersection simultaneously, blocking each other's paths and causing a complete stop of traffic. To recreate this in CARLA, we set up a controlled intersection where several vehicles from different lanes approach the intersection but fail to clear it due to conflicting movements, as show in Fig~\ref{fig6}.
 In this case, the VLM identifies the main cause of the congestion as multiple vehicles blocking the intersection, failing to adhere to the right-of-way. The system then suggests specific actions for each vehicle: For instance, vehicle v-8 is instructed to move backward 5 meters at increasing speed, vehicle v-0 is directed to move backward by 6 meters, vehicle v-3 is instructed to move backward 5 meters, and vehicle v-6 is directed to move forward while maintaining speed. The effectiveness of the solution is evaluated based on how quickly the intersection is cleared and how efficiently normal traffic flow is restored.

\begin{figure}[h]
      \centering
    \includegraphics[width=0.5\textwidth, height=0.55\textwidth, keepaspectratio]{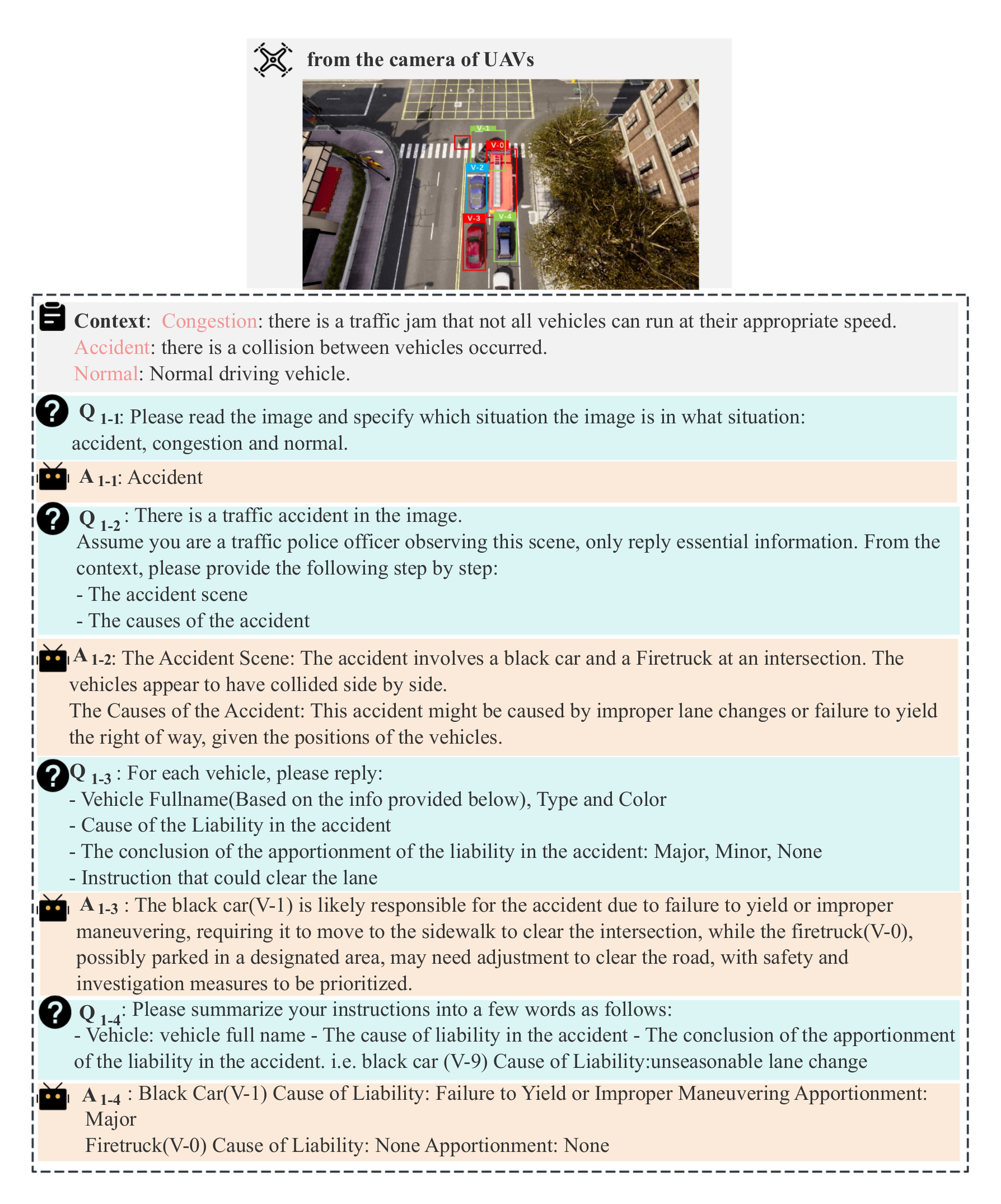}
    \vspace{-15pt} 
      \caption{the process of VLM handling accident}
      \label{fig7}
      \vspace{-10pt} 
    \end{figure}

\textbf{Accident Responsibility Allocation:} The third experiment simulates a traffic accident involving two vehicles at an intersection, where the responsibility for the accident needs to be assigned, and the involved vehicles must be relocated to prevent further accidents. In this scenario, CARLA simulates a collision at a busy intersection, with a black car and a firetruck involved in the crash, blocking traffic, as show in Fig~\ref{fig7}.
 In this accident scene, where the black car (v-9) and the firetruck (v-0) have collided. It determines that the accident was likely caused by the black car's failure to yield or improper lane change, according to traffic laws and vehicle positions. The VLM assigns responsibility, stating that the black car is primarily at fault for failing to yield, while the firetruck is not liable for the incident. 
\subsection{Time Overhead}
We tested the time overhead of ChatGPT-4o for three traffic anomalies, with each data point being tested 20 times and the average value recorded in Table~\ref{table:time_overhead}. As observed, the Scene Stage consistently exceeds 8 seconds, as this stage involves uploading and processing images, which requires significant computational effort. Subsequently, the time overhead gradually decreases across the remaining stages, with the final stage being the most efficient, as it primarily focuses on data reorganization and command generation, requiring minimal processing. Overall, all three scenarios are completed in approximately 14 seconds, demonstrating a significantly higher efficiency compared to human intervention on-site, which typically involves manual assessment, decision-making delays, and communication overhead. This efficiency highlights the potential of VLM-powered automation in real-time traffic anomaly resolution.
    \begin{table}[!tb]
    \vspace{10pt}
    \centering
    \caption{ChatGPT-4o Time Overhead}
    \label{table:time_overhead}
    \setlength{\tabcolsep}{2pt} 
    \begin{tabular}{|c|c|c|c|c|c|}
        \hline
        \begin{tabular}{c} Scene \\ Name \end{tabular} & \begin{tabular}{c} Scene \\ Stage \end{tabular} &\begin{tabular}{c}Analysis \\ Stage \end{tabular} &\begin{tabular}{c}Solution \\ Stage \end{tabular} &\begin{tabular}{c}Formatting \\ Stage\end{tabular} & Total\\
        \hline
        \begin{tabular}{c} Ghost \\ Traffic Jam \end{tabular} &8.2s &1.9s &2.0s &1.1s &13.2s\\
        \hline
        \begin{tabular}{c} Intersection \\ Deadlock \end{tabular} &9.1s &2.0s &2.3s &1.4s &14.8s\\
        \hline
        Accident &8.5s &1.7s &2.2s &1.4s &13.6s\\
        \hline
    \end{tabular}
    \vspace{-10pt}
    \end{table}  
\section{Conclusion}
In this paper, we proposed a closed-loop testing framework using VLMs in the CARLA simulator to address traffic anomalies. By replicating critical scenarios such as phantom traffic jams, intersection deadlocks, and collisions, our framework demonstrates VLMs' ability to analyze and resolve anomalies in real time. Integrating a novel CoT approach, we enable seamless translation of VLM-generated solutions into executable commands, showcasing their potential for autonomous traffic management. Our results validate the feasibility of VLMs in enhancing traffic anomaly resolution and lay the foundation for their integration into scalable urban traffic systems. Future work will focus on handling more complex scenarios and evaluating real-world effectiveness. Additionally, we will continue to expand the range of traffic anomaly scenarios in our test tool, while also exploring the capabilities of other VLMs to further reduce time overhead and improve processing efficiency.

\end{document}